\documentclass[conference]{IEEEtran}
\IEEEoverridecommandlockouts
\usepackage{cite}
\usepackage{amsmath,amssymb,amsfonts}
\usepackage{algorithmic}
\usepackage{graphicx}
\usepackage{textcomp}
\usepackage{xcolor}
\usepackage{flushend}
\usepackage{enumitem, url}
\usepackage{subcaption}
\usepackage{booktabs}
\usepackage{url}
\usepackage{multicol}
\usepackage{tikz}
\usetikzlibrary{shapes.geometric, arrows, positioning, fit}
\usepackage{multirow}
\def\BibTeX{{\rm B\kern-.05em{\sc i\kern-.025em b}\kern-.08em
    T\kern-.1667em\lower.7ex\hbox{E}\kern-.125emX}}

\title{Mamba-Based Ensemble learning for White Blood Cell Classification}

\author{
Lewis Clifton$^{1}$, Xin Tian$^{2}$, Duangdao Palasuwan$^{3}$, Phandee Watanaboonyongcharoen$^{4}$, \\
Ponlapat Rojnuckarin$^{5}$, Nantheera Anantrasirichai$^{6}$ \\
\small
$^{1}$Computer Science, University of Bristol, UK, js21767@bristol.ac.uk \\
$^{2}$Centre for Human Genetics, University of Oxford, UK, xin.tian@well.ox.ac.uk \\
$^{3}$Oxidation in Red Cell Disorders Research Unit, Faculty of Allied Health Sciences, Chulalongkorn University, Thailand, nantadao@gmail.com \\
$^{4}$Department of Laboratory Medicine, Faculty of Medicine, Chulalongkorn University, Thailand, phandee\_lee@yahoo.com \\
$^{5}$Excellence Center in Translational Hematology, Faculty of Medicine, Chulalongkorn University, Thailand, Ponlapat.R@Chula.ac.th \\
$^{6}$Visual Information Laboratory, University of Bristol, UK, N.Anantrasirichai@bristol.ac.uk
}

\begin{document}

\maketitle

\begin{abstract}

White blood cell (WBC) classification assists in assessing immune health and diagnosing various diseases, yet manual classification is labor-intensive and prone to inconsistencies. Recent advancements in deep learning have shown promise over traditional methods; however, challenges such as data imbalance and the computational demands of modern technologies, such as Transformer-based models which do not scale well with input size, limit their practical application. This paper introduces a novel framework that leverages Mamba models integrated with ensemble learning to improve WBC classification. Mamba models, known for their linear complexity, provide a scalable alternative to Transformer-based approaches, making them suitable for deployment in resource-constrained environments. Additionally, we introduce a new WBC dataset, Chula-WBC-8, for benchmarking. Our approach not only validates the effectiveness of Mamba models in this domain but also demonstrates their potential to significantly enhance classification efficiency without compromising accuracy. The source code can be found at \url{https://github.com/LewisClifton/Mamba-WBC-Classification}.

\end{abstract}

\begin{IEEEkeywords}
Image classification, white blood cell, microscopic image, ensemble learning, Mamba, Transformer
\end{IEEEkeywords}

\section{Introduction}
\label{sec:intro}

Classifying white blood cell (WBC) types is crucial, primarily because different types of WBCs serve distinct functions in the immune response. Accurate classification provides insights into an individual's immune health, aids in diagnosing infections and diseases, and guides treatment. 

Manual WBC classification is labor-intensive and requires specialized training, often leading to diagnostic inconsistencies due to the subjectivity of blood morphology and human factors such as fatigue or distraction~\cite{automation_challenges_fuentes}. Although automated systems integrated into blood imaging analyzers aim to improve efficiency, their accuracy is questionable; some WBC types are detected correctly less than 50\% of the time, and the systems (e.g. DI-60) are costly~\cite{sysmex_zhao}. Anecdotal and empirical evidence points to significant limitations of the DI-60 system, despite its contributions to workflow efficiency~\cite{sysmex_zhao, sysmex_kim, sysmex_tabe, sysmex_nam}. Notably, the DI-60 struggles with accurately classifying WBC abnormalities crucial for diagnosing conditions like leukemia~\cite{sysmex_zhao}. Furthermore, while the DI-60's capabilities can support workflows in a supplementary role, they fall short without expert oversight~\cite{sysmex_kim,sysmex_tabe}, ultimately increasing turnaround times, contradicting the intended benefits~\cite{sysmex_nam}.

\begin{figure}[t]
    \centering
    \includegraphics[width=\linewidth]{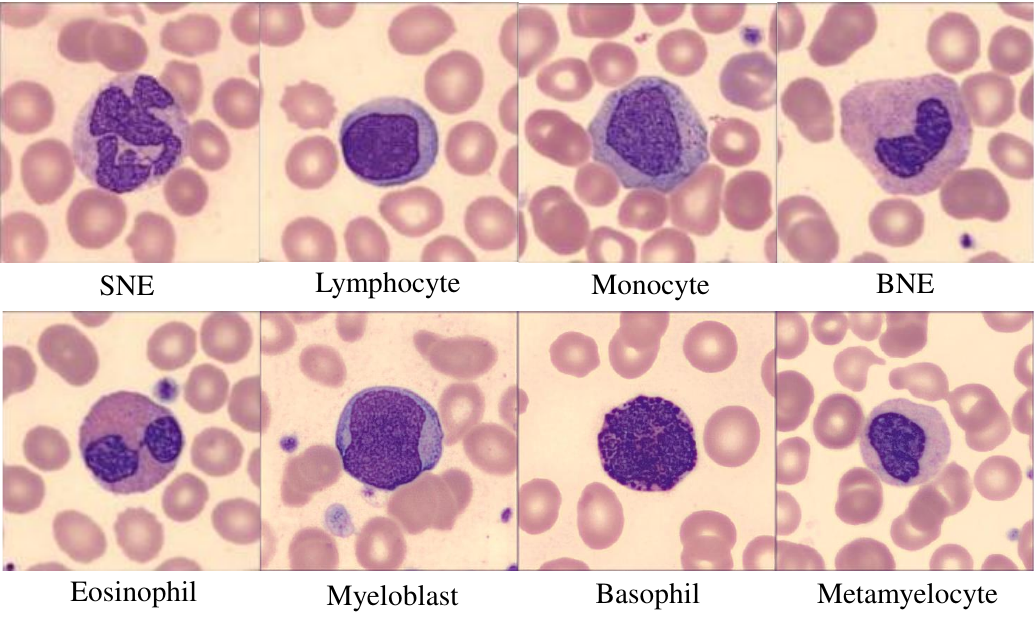} 
    \caption{Example images from the Chula-WBC-8 dataset of its eight WBC types: segmented neutrophils (SNE), lymphocytes, monocytes, band neutrophils (BNE), eosinophils, myeloblasts, basophils, and metamyelocytes in peripheral blood smears.} 
    \label{fig:chula_wbc}
    \vspace{-4mm}
\end{figure}


The emergence and development of new technology provide opportunities to address these limitations. Advances in Deep Learning (DL) have significantly enhanced WBC counting and differential classification, surpassing traditional approaches~\cite{automation_challenges_khamael, automation_challenges_shah}. More recently, Transformer architectures have been shown to be effective backbones for vision tasks. However, the attention mechanism attributed to the models' success suffers from quadratic complexity with respect to the input size, making them computationally expensive and difficult to deploy on hardware-constrained systems common in clinical laboratories~\cite{automation_challenges_han}.

This led to the proposal of Mamba~\cite{mamba}, a model offering linear complexity as a scalable alternative in Transformer-dominated domains. Recent studies have demonstrated the effectiveness of Mamba-based models across various vision tasks~\cite{mambavision, vmamba, vim, localmamba, plainmamba, medmamba, survey_bansal, survey_ibrahim, survey_liu, survey_qu, survey_rahman, survey_xu, survey_zhang}; however, the potential of Mamba remains unexplored for WBC classification.


 Current DL based WBC classification suffers from several challenges in WBC classification such as dataset availability, size and imbalance~\cite{automation_challenges_asghar}. This arises due to the natural imbalance WBC subtypes in the blood and is a common problem among WBC datasets. Data imbalance causes bias towards majority classes, producing worse performance for rarer subtypes.
 
This paper proposes a novel framework based on Mamba models and their integration with ensemble learning for WBC classification. We also introduce a new WBC dataset (Fig.~\ref{fig:chula_wbc}) to evaluate these models and address the emergent imbalance between WBC subtypes. The subsequent discussion of these results reveals the validity of Mamba as a tool in this domain and its potential to improve the efficiency of classification systems with minimal compromise to predictive accuracy. The main contributions of this paper can be summarized as follows: 
\begin{itemize}[noitemsep,topsep=0pt,leftmargin=10pt, itemindent=0pt]
\item This paper presents the \textit{first} use of Mamba models for WBC classification.
\item Our novel framework is based on ensemble learning and integrates techniques for handling data imbalance.
\item A new dataset, Chula-WBC-8, comprises eight types of WBCs from patients with WBC disorders.
\item Comprehensive experiments comparing with Sysmex DI-60 and state-of-the-art deep learning approaches are conducted.
\end{itemize}

\section{Related Work and Background}
\label{sec:related}

\subsection{Existing white blood cell classification methods}

The initial application of DL for WBC classification utilized transfer learning, adapting models pre-trained on natural images using architectures such as VGG \cite{Ozyurt2020WBCDetection}, ResNet, and Inception \cite{Habibzadeh2017}. Later, advancements in the field introduced more sophisticated CNN-based models, such as DenseNet121 \cite{DeepLearningWBC2023} and YOLO \cite{Abas2022YOLOLeukocytes} to WBC classification. These architectures were specifically chosen for their capability to capture detailed and nuanced features of WBCs. Recent developments have integrated Transformer models and self-attention mechanisms, which excel in processing both local and global features within images \cite{Katar2023WBCViT}. This approach allows for more dynamic and context-aware feature analysis across the spatial domain of the images. Additionally, the integration of transformer-based object detection, such as DETR \cite{LENG2023104518}, introduces end-to-end training capabilities, significantly improving the precision and reliability of WBC classification.

\subsection{Mamba}

State Space Models (SSMs) map input sequences to output sequences through a latent state calculated from the previous state and current input, offering an efficient alternative to Transformers by scaling linearly with sequence length and enabling parallel computation. The Mamba frameworks utilize Structured State Space sequence model (S4) within a DL framework to manage long sequence states effectively. Mamba's selective scan mechanism updates parameters to determine which features of input tokens propagate forward. Its causal convolution blocks perform convolutions that are appropriate for sequences, reducing spatial dimensions of the input while ensuring that operations are limited to previous states. For visual data, input images are divided into spatially encoded patches fed into a stack of Mamba blocks, which progressively down-sample feature maps to capture longer sequential dependencies in deeper blocks.

Mamba models for imaging tasks include several variants with distinct enhancements. Vision Mamba (Vim)\cite{vim} refines the standard Mamba block with a bidirectional approach and an extra branch for reverse convolution and reverse SSM networks to enhance global context representation. VMamba\cite{vmamba} applies Mamba to vision tasks using a novel 2D selective scan module that processes image patches from diverse scanning paths to generate detailed 2D feature maps. MambaVision~\cite{mambavision} swaps causal convolution for standard convolution to allow richer context consideration by integrating both sequential and spatial dependencies. PlainMamba~\cite{plainmamba} maintains spatial adjacency and semantic continuity through a bi-directional scan path and directional encoding. LocalMamba~\cite{localmamba} captures intra-region dependencies with local scans and a dual-branch attention mechanism for managing global and local feature interactions. MedMamba~\cite{medmamba} incorporates grouped convolution for efficient feature capturing, showing high accuracy in tests like BloodMNIST \cite{medmnist} but remains underexplored for specific applications such as WBC classification, highlighting a significant area for future research \cite{survey_bansal}.

\section{Datasets and Preparation}
\label{sec:dataset}
This paper uses two datasets: BloodMNIST, an accessible benchmark for normal blood cell classification, and a newly introduced dataset, Chula-WBC-8, which contains cells from WBC disorders. The latter presents a greater challenge for automated classification.

\subsection{BloodMNIST}
BloodMNIST, part of the MedMNIST dataset~\cite{medmnistv2}, comprises 17,092 microscopic images of normal cells from healthy individuals free of infections, hematologic or oncologic diseases, and pharmacological treatments at the time of collection. The dataset is organized into 8 classes: neutrophils, eosinophils, basophils, lymphocytes, monocytes, immature granulocytes (promyelocytes, myelocytes, and metamyelocytes), erythroblasts and platelets or thrombocytes. As per the original paper, it is pre-partitioned into training, validation, and test sets in a 7:1:2 ratio. Initially, the images have a resolution of 360$\times$363 pixels; they are center-cropped to 200$\times$200 and subsequently resized to 28$\times$28 pixels.

\subsection{Chula-WBC-8}
We collected 4,808 images of peripheral blood smears (Table~\ref{tab:class_frequencies_augmented}) using a Sysmex DI-60 machine from a sample group of 113 patients with WBC disorders at Chulalongkorn Hospital, Thai Red Cross Society. The types of WBCs from the peripheral blood smears were identified by five experts in a blind test, with the majority opinion, which is three out of five, used to confirm the answers. All WBC images were anonymized to ensure patient privacy. Table~\ref{tab:class_frequencies_augmented} shows the class frequencies for the dataset. The dataset exhibits significant inter-class imbalance, a common issue in WBC datasets due to the natural variation in WBC subtype prevalence.

\begin{table}[t]
    \centering
    \setlength{\tabcolsep}{4pt}
    \caption{Chula-WBC-8 dataset with class frequencies and augmentation details}
    \label{tab:class_frequencies_augmented}
    \begin{tabular}{lccc}
        \toprule
        Class & Original & Augmented & Total \\
        \midrule
        SNE                    & 1985  & 0   & 1985 \\
        Lymphocyte             & 1253  & 0   & 1253 \\
        Monocyte               & 567   & 0   & 567  \\
        BNE                    & 514   & 86  & 500  \\
        Eosinophil             & 157   & 343 & 500  \\
        Myeloblast             & 156   & 344 & 500  \\
        Basophil               & 93    & 407 & 500  \\
        Metamyelocyte          & 83    & 417 & 500 \\
        \midrule
        \textbf{Total:} & 4808 & 1607 & 6215 \\
        \bottomrule
    \end{tabular}
    \vspace{-3mm}
\end{table}

\section{Methodology}
\label{sec:method}

\begin{figure}[t]
\begin{center}
   \includegraphics[width=\linewidth]{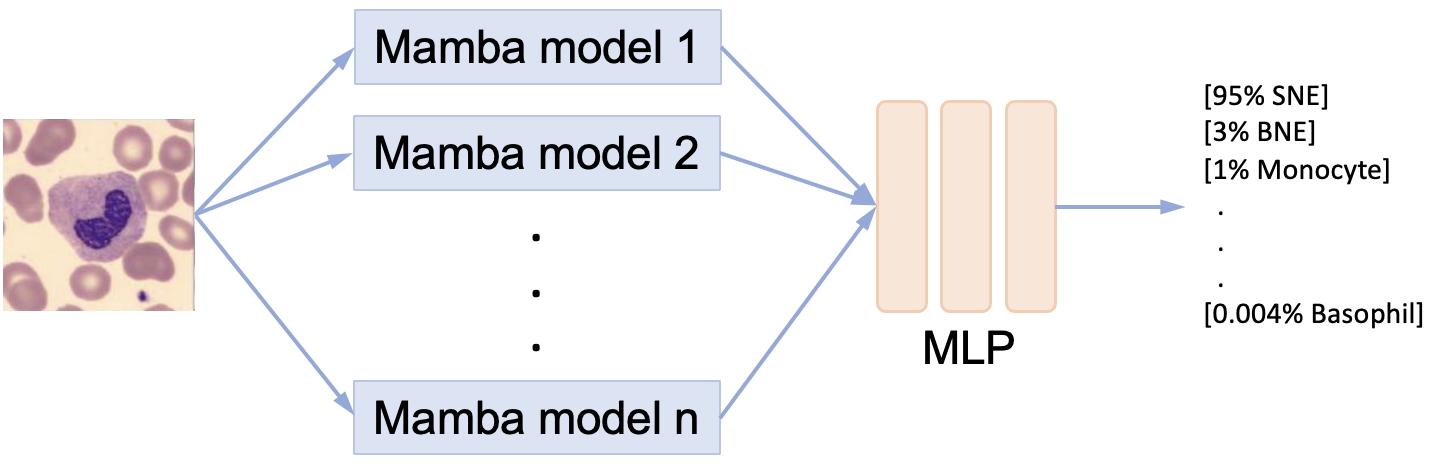}
\end{center}
   \caption{Diagram of the proposed Mamba-based ensemble learning framework (n=5 in this paper). }
\label{fig:diagram}
\vspace{-3mm}
\end{figure}

The proposed diagram is shown in Fig.~\ref{fig:diagram}, combining five Mamba-based models using ensemble learning.
\subsection{Mamba architectures}

Five Mamba-based methods are employed in our study, including Vision Mamba (ViM), VMamba~\cite{vmamba}, MambaVision~\cite{mambavision}, MedMamba~\cite{medmamba}, and LocalMamba~\cite{localmamba}.

\subsubsection{Vision Mamba (ViM)~\cite{vim}} A bi-directional version of the standard Mamba block is employed with the introduction of a third branch that uses reverse convolution and reverse SSM networks. This addition gives ViM a stronger representation of the global context of the input.

\subsubsection{VMamba~\cite{vmamba}} A novel 2D selective scan (SS2D) module is proposed. It forms four sequences of image patches obtained by traversing different scanning paths over the input patches. These are fed into independent S6 blocks (selective SSM) and cross-merged to produce a 2D feature map. The stacking of SS2D produces more granular feature maps deeper in the network, capturing finer details.

\subsubsection{MambaVision~\cite{mambavision}} swaps the causal convolution layer used by conventional Mamba with standard convolution to remove the restriction of considering the state sequence in one direction, which is inappropriate given the spatial nature of images. A second convolution layer is used in the secondary path,developing a richer global context by considering both sequential and spatial dependencies of the input in both paths of the block.

\subsubsection{LocalMamba~\cite{localmamba}} Local scans in four parallel directions capture intra-region dependencies and global interactions, employing a dual-branch attention mechanism. The spatial branch aggregates global information, while the channel branch adjusts token importance. This method captures detailed local and global features and dynamically optimizes scan paths for each layer, ensuring efficient feature prioritization and comprehensive interaction analysis.

\subsubsection{MedMamba~\cite{medmamba}} The modified VMamba block incorporates a grouped convolution layer in its secondary branch to efficiently capture a broad spectrum of features. Grouped convolutions require a channel-shuffle layer to facilitate feature integration and prevent information loss between channels. This setup enhances feature representation by combining sequential and spatial dependencies.


\subsection{Ensemble learning}

Ensembles benefit from leveraging diverse base models to maximize variance, combining the predictions of several trained models through a ``meta-learner" to enhance performance. In this paper, our training set is partitioned into two subsets: one for training base Mamba models and the other as a holdout set for training the meta-learner, preventing data leakage and ensuring that the training metrics for the meta-learner remain optimistic.
While training the meta-learner, the base models generate predictions on the holdout data; these predictions are then concatenated as input for the meta-learner. The meta-learner is a simple Multilayer Perceptron (MLP) trained for 5 epochs to avoid overfitting, optimizing its ability to effectively integrate insights from the base models without inheriting their biases or specific errors.

\subsection{Dealing with imbalanced dataset}

We employ data augmentation and a weighted loss function to deal with imbalanced dataset.

 The data augmentation applied includes random rotation (0° to 360°), translation (±10\% along both x and y axes), scaling (90\% to 110\%), and shearing (±5°). Horizontal and vertical flips are performed with a 50\% probability each. Brightness and contrast are adjusted within ±10\%, saturation within ±5\%, and hue within ±2\%. Additionally, Gaussian blur is applied with a kernel size of 3 and sigma ranging from 0.1 to 1.0. Table~\ref{tab:class_frequencies_augmented} shows the resulting class frequencies following data augmentation for Chula-WBC-8 dataset. 

In initial training phase, the original models utilize a cross-entropy loss function, which is standard for classification tasks. To further address class imbalance, we employ a weighted cross-entropy loss, assigning weights inversely proportional to the class frequencies, thus emphasizing minority classes.

\subsection{Training settings}

All models are trained using the AdamW optimizer. The training configuration for each model follows the settings specified in the original papers. The training was done using a single Nvidia RTX 4070 GPU.

The datasets used in this study are divided into training and test splits. The BloodMNIST dataset follows its predefined split, with 11,959 images for training, 1712 for validation, and 3,421 for testing. The Chula-WBC-8 dataset is partitioned using a 4:1 ratio for training and testing, respectively. Images are resized to \(224\times224\) and normalized prior to being fed to the model.

\section{Experimental results and discussion}
\label{sec:results}

\subsection{Evaluation Metrics}
Model performance is evaluated using standard classification metrics: Accuracy (Acc.), Precision (Prec.), Sensitivity (Sens.), and F1 Score (F1). Accuracy quantifies the proportion of correctly classified samples, Precision indicates the fraction of predicted positives that are correct, Sensitivity measures the fraction of actual positives identified, and F1 provides their harmonic mean. Given the class imbalance, we report weighted averages for all metrics.

\subsection{Results of BloodMNIST dataset}

Table~\ref{tab:BloodMNIST} presents the results of WBC classification. Mamba models outperform Google AutoML, the top benchmark in \cite{medmnistv2} by  2.7\%. The BloodMNIST dataset does not display a significantly skewed distribution, resulting in minimal differences in performance between models trained with and without an imbalance strategy. Notably, Medmamba produced the poorest results, likely because we did not have access to the pretrained model, necessitating training from scratch. In contrast, LocalMamba and MambaVision are the top performers among individual models. Our ensemble learning approach did not yield significant improvement, as the overall performance on the BloodMNIST dataset was already high, indicating that all models are generally effective.

\begin{table}[t]
    \centering
    \caption{Performance of the models on BloodMNIST. \textbf{Bold} and \underline{underline} indicate the best and second best performers.}
    \begin{tabular}{lcccc}
        \toprule
        Method & Acc. & F1 & Prec. & Sens  \\
        \midrule
        ResNet-50~\cite{medmnistv2}  & 95.0\% & - & - & - \\
        Google AutoML~\cite{medmnistv2} & 96.6\% & - & - & - \\
        \midrule
        \multicolumn{5}{c}{\textbf{Without Imbalanced Data Strategy}} \\ \midrule
        SWIN ViT \cite{swin}            & 98.80\% & 0.9880 & 0.9880 & 0.9880   \\
        VMamba~\cite{vmamba}            & 98.91\% & 0.9891 & 0.9892 & 0.9892  \\
        Vim~\cite{vim}                  & 98.45\% & 0.9845 & 0.9847 & 0.9845  \\
        Medmamba~\cite{medmamba}        & 94.48\% & 0.9446 & 0.9490 & 0.9448  \\
        MambaVision~\cite{mambavision}  & 98.80\% & 0.9880 & 0.9880 & 0.9880 \\
        LocalMamba~\cite{localmamba}    & \underline{99.04}\% & \underline{0.9904} & \underline{0.9904} & \underline{0.9904}  \\
        Ensemble(ours)      & \textbf{99.24}\% & \textbf{0.9925} & \textbf{0.9924} & \textbf{0.9924} \\
        \midrule
        \multicolumn{5}{c}{\textbf{With Imbalanced Data Strategy}} \\ \midrule
        SWIN ViT \cite{swin}                &  98.42\% & 0.9843 & 0.9850 & 0.9842   \\
        VMamba~\cite{vmamba}                & 97.75\% & 0.9776 & 0.9781 & 0.9775   \\
        Vim~\cite{vim}                      & 98.71\% & 0.9871 & 0.9872 & 0.9871  \\
        Medmamba~\cite{medmamba}            & 91.14\% & 0.9081 & 0.9299 & 0.9114  \\
        MambaVision~\cite{mambavision}      & \underline{98.91}\% & \underline{0.9892} & \underline{0.9892} & \underline{0.9892}  \\
        LocalMamba~\cite{localmamba}        & 98.36\% & 0.9836 & 0.9838 & 0.9836   \\
        Ensemble(ours)      & \textbf{99.12}\% & \textbf{0.9912} & \textbf{0.9914} & \textbf{0.9912} \\
        \bottomrule
    \end{tabular} 
    \label{tab:BloodMNIST}
    \vspace{-2mm}
\end{table}

\subsection{Results of Chula-WBC-8 dataset}

Table~\ref{tab:Chula} shows the results of Sysmex DI-60 WBC classification, and DL models both with and without adjustments for data imbalance. Given the imbalanced nature of this dataset, improvements are evident when strategies to address imbalance are employed in most models, except for the ensemble model. This might be that ensembling could naturally overcome the imbalance issue. The Mamba models outperform DI-60 by up to 9.6\%. 

\begin{figure}[t]
    \centering
\includegraphics[width=\linewidth]{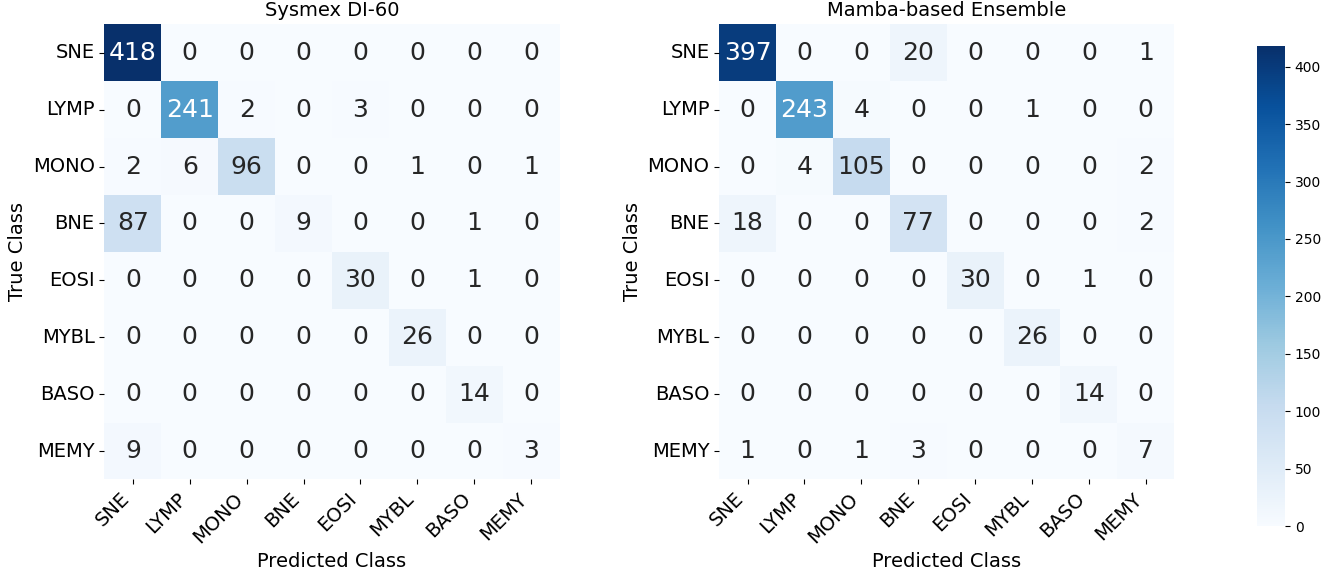} 
    
    \caption{Comparison of confusion matrices: Sysmex DI-60 (left) vs. Mamba-based ensemble learning (right).}
    \label{fig:confusionmatrix}
    \vspace{-2mm}
\end{figure}

Fig.~\ref{fig:confusionmatrix} compares the confusion matrix results from the Sysmex DI-60 and our Mamba-based ensemble learning approach. Both methods predominantly misclassify between SNE (segmented neutrophils) and BNE (band neutrophils), often due to some cells being in a transitional stage from BNE to SNE, resulting in ambiguous classifications. This comparison highlights both the challenges and the effectiveness of our model in handling classes with overlapping characteristics. 

\begin{figure}[t]
    \centering
    



    \includegraphics[width=0.95\linewidth]{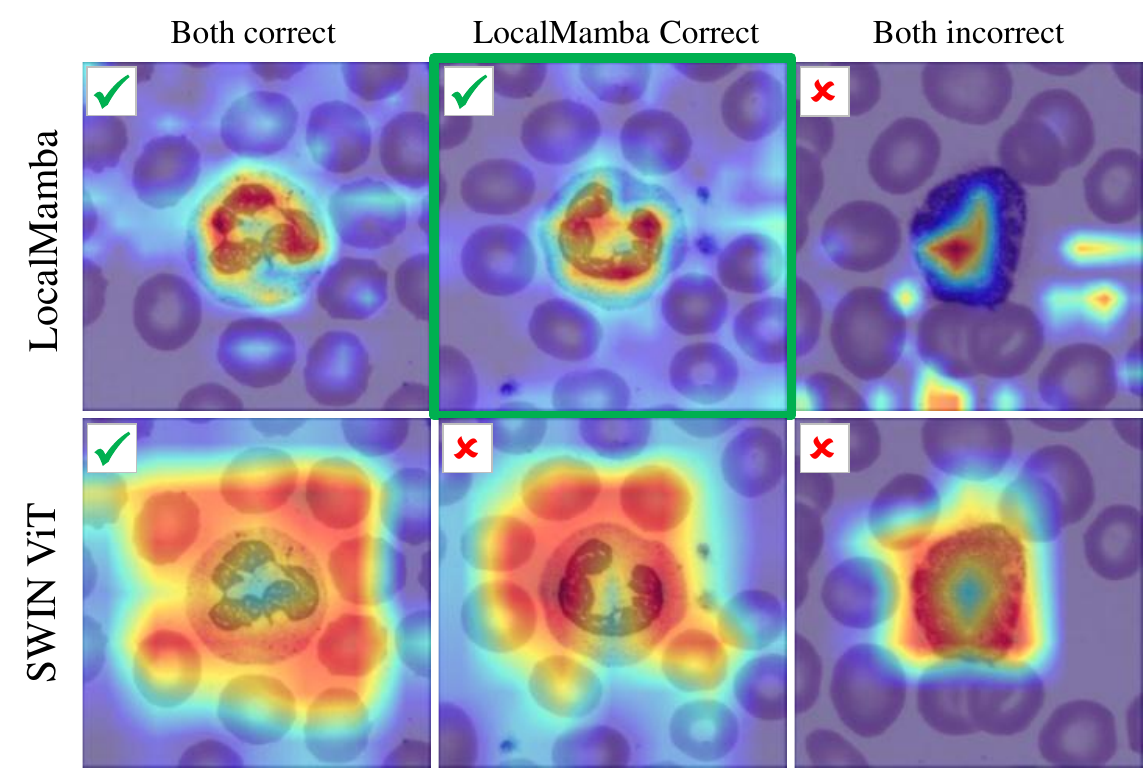} 
\caption{Heatmap visualizations of key features (red indicates higher importance) across scenarios: (Left) Both correct, (Middle) Only LocalMamba correct, (Right) Both incorrect. Rows: LocalMamba (top), Swin Transformer (bottom).}
    \label{fig:gradcam_comparison}
    \vspace{-3mm}
\end{figure}

Fig.~\ref{fig:gradcam_comparison} shows the heatmap where key features are used for decision-making. Mamba focuses more on the cell area than Swin Transformer. 

Overall performance is lower on Chula than on BloodMNIST because Chula has lower inter-class variance, making it more difficult to distinguish between classes. Chula contains only WBCs, whereas BloodMNIST includes WBCs, RBCs, and platelets, leading to greater variation in the dataset.

\begin{table}[t]
    \centering
    \caption{Performance of the models on Chula-WBC-8. \textbf{Bold} and \underline{underline} indicate the best and second best performers.}
    \begin{tabular}{lcccc}
        \toprule
        Method & Acc. & F1 & Prec. & Sens  \\ \midrule
        Sysmex DI-60 & 85.68\% & 0.9013 & 0.9720 & 0.8568 \\ \midrule  
        \multicolumn{5}{c}{\textbf{Without Imbalanced Data Strategy}} \\ \midrule
        SWIN ViT \cite{swin}               & \underline{92.69}\% & \underline{0.9265} & \underline{0.9263} & \underline{0.9269} \\
        VMamba~\cite{vmamba}                & 91.95\% & 0.9212 & 0.9241 & 0.9195 \\
        Vim~\cite{vim}                      & 88.71\% & 0.8938 & 0.9092 & 0.8871 \\
        Medmamba~\cite{medmamba}            & 65.83\% & 0.7625 & 0.6583 & 0.6684  \\
        MambaVision~\cite{mambavision}      & 90.80\% & 0.9126 & 0.9080 & 0.9097 \\
        LocalMamba~\cite{localmamba}        & 90.80\% & 0.9170 & 0.9080 & 0.9105  \\
        {Ensemble (ours)}      & \textbf{93.94}\% & \textbf{0.9397} & \textbf{0.9401} & \textbf{0.9393} \\
        \midrule
        \multicolumn{5}{c}{\textbf{With Imbalanced Data Strategy}} \\ \midrule
        SWIN ViT \cite{swin}               & 91.12\% & 0.9087 & 0.9114 & 0.9112 \\
        VMamba~\cite{vmamba}               & 87.15\% & 0.8824 & 0.9190 & 0.8715  \\
        Vim~\cite{vim}                      & 89.66\% & 0.8921 & 0.8949 & 0.8966 \\
        Medmamba~\cite{medmamba}            & 87.04\% & 0.8454 & 0.8533 & 0.8704  \\
        MambaVision~\cite{mambavision}      & 92.00\% & 0.9186 & 0.9192 & 0.9192  \\
        LocalMamba~\cite{localmamba}        & \textbf{92.79}\% & \textbf{0.9272} & \textbf{0.9269} & \textbf{0.9279}  \\
        Ensemble  (ours)      & \underline{92.37}\% & \underline{0.9202} & \underline{0.9199} & \underline{0.9237} \\
        
        \bottomrule
    \end{tabular}
    \label{tab:Chula}
    \vspace{-2mm}
\end{table}

\section{Conclusion}
\label{sec:conclusion}

This paper presents a new framework for WBC classification. We employ modern architectures, specifically Mamba, and leverage ensemble learning to enhance overall performance. Additionally, we introduce a new dataset containing eight types of WBCs from patients with disorders. To address dataset imbalance, we apply data augmentation and a weighted loss function. The results demonstrate that our method outperforms CNNs, Swin Transformer and the automated software provided by Sysmex DI-60.

 Future research could consider these downstream tasks, supplementing our own work to form a comprehensive overview of the potential of Mamba-based models in WBC tasks as well as potentially enabling an alternate approach to WBC-based diagnostics, such as automated Leukaemia detection.

\bibliographystyle{IEEEtran}
\bibliography{bib}

\end{document}